\documentclass[preprint,12pt,3p]{elsarticle}
\usepackage{colortbl}
\usepackage{graphicx}
\usepackage{epstopdf}
\usepackage{amsmath,bm}
\usepackage{amsthm}
\usepackage{amssymb}
\usepackage{amsfonts}
\usepackage{algorithmic}
\usepackage{algorithm}
\usepackage{array}
\usepackage{multirow}
\usepackage{booktabs}
\usepackage{color}
\usepackage{float}
\usepackage{subfig}
\usepackage{makecell}
\usepackage{threeparttable} 
\usepackage{cases}
\usepackage{ragged2e}
\usepackage[colorlinks]{hyperref}
\usepackage{xr}
\usepackage{soul}
\usepackage[switch,pagewise]{lineno}
\journal{arXiv}
\begin{document}

\begin{frontmatter}

\title{PatchProto Networks for Few-shot Visual Anomaly Classification}

\author[label1]{Jian Wang}
\address[label1]{College of Control Science and Engineering, Zhejiang University}

\ead{wangjianxxx818@gmail.com}

\author[label1]{Yue Zhuo\corref{cor1}}
\cortext[cor1]{Corresponding author}
\ead{zhuoy1995@zju.edu.cn}

\begin{abstract}
The visual anomaly diagnosis can automatically analyze the defective products, which has been widely applied in industrial quality inspection. The anomaly classification can classify the defective products into different categories. However, the anomaly samples are hard to access in practice, which impedes the training of canonical machine learning models. This paper studies a practical issue that anomaly samples for training are extremely scarce, i.e., few-shot learning (FSL). Utilizing the sufficient normal samples, we propose PatchProto networks for few-shot anomaly classification. Different from classical FSL methods, PatchProto networks only extract CNN features of defective regions of interest, which serves as the prototypes for few-shot learning. Compared with basic few-shot classifier, the experiment results on MVTec-AD dataset show PatchProto networks significantly improve the few-shot anomaly classification accuracy.

\end{abstract}


\end{frontmatter}

\section{Introduction}
\label{intro}
\par The anomaly diagnosis based on computer-vision techniques can automatically analyze the defective samples, including tasks like anomaly detection, segmentation and classification. Anomaly detection~\cite{10.1145/3439950,ma2021comprehensive} has been a prevalent research topic recently, which aims to identify the unusual patterns (i.e., outliers) that deviate from the normal observations. Anomaly segmentation~\cite{Yi_2020_ACCV,SHI20219} highlights the anomaly variables (e.g., pixels) in input space, which provides the explanation for the detection results. Furthermore, anomaly classification seeks to recognize what types the anomaly samples belong to, which could benefit the maintenance of the industrial systems.

\par The major impediment of anomaly classification is the scarcity of anomaly samples. This issue recently has been an active research area in the flied of machine learning, also known as few-shot learning (FSL). FSL classifies new data with only a few training samples\footnote{When there is only one training sample of each class, the problem becomes one-shot learning.} with supervised information. Anomaly classification is a typical example of FSL, since the defective samples rarely occur and are hard to collect. In past few years, many FSL approaches are proposed, including classical models like Model-Agnostic Meta-Learning (MAML)~\cite{DBLP:journals/corr/FinnAL17}, Matching Networks~\cite{DBLP:journals/corr/VinyalsBLKW16}, and Prototypical Networks~\cite{DBLP:journals/corr/SnellSZ17}.

\par Industrial anomaly classification is a little different. Unlike the universal FSL benchmarks (e.g., Omniglot~\cite{Omniglot} and CUB~\cite{2011The}) consisting of hundreds of class labels, the number of industrial anomaly type is relatively small, thus it is commonly infeasible to apply meta-training in FSL anomaly classification. Fortunately, industrial datasets contains a large number of normal samples, which can be utilized for classifying the anomaly samples.

\par
\begin{figure*}[t]
  \centering
  \includegraphics[width=0.95\textwidth]{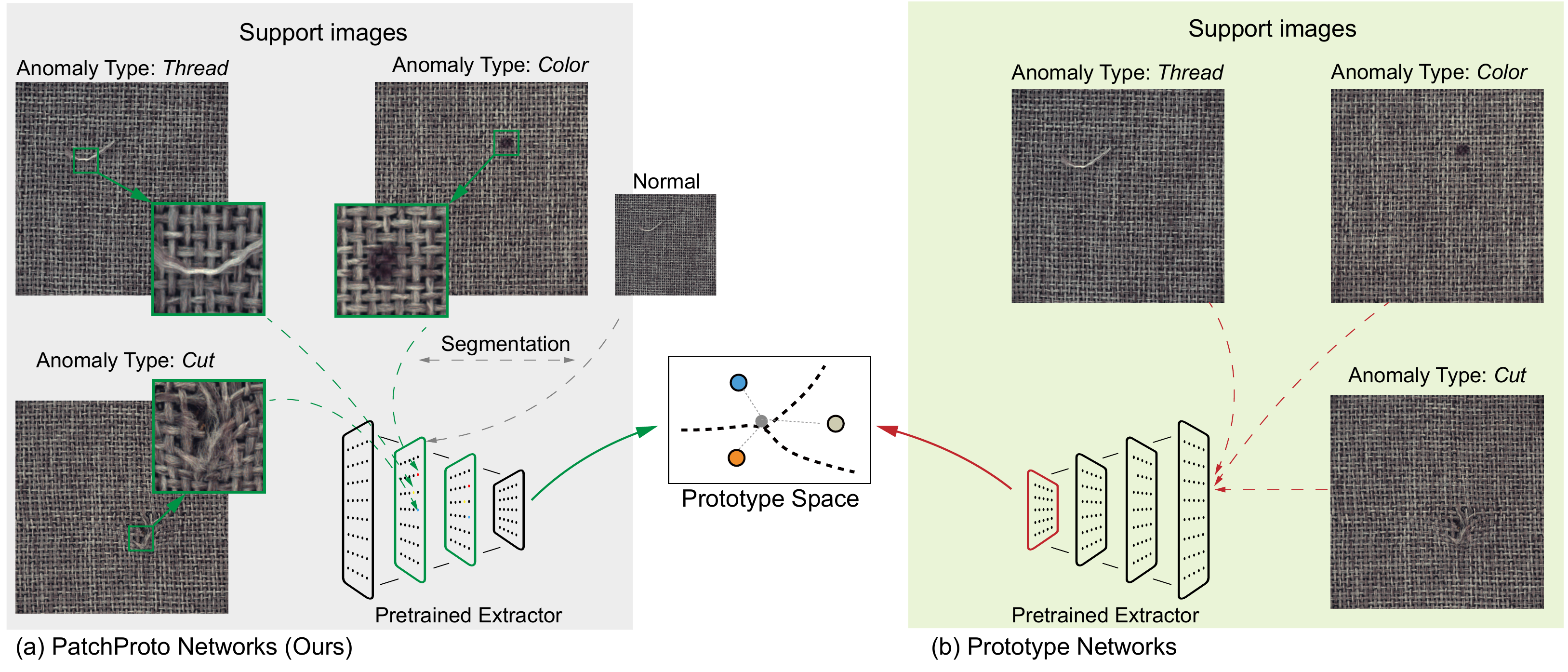}
  \caption{\textbf{Comparison between (a) PatchProto and (b) Prototype networks} for textile anomaly classification in MVTec-AD dataset~\cite{Bergmann_2019_CVPR}. }
  \label{Paths_comparison}
\end{figure*}

 Inspired by the anomaly detection methods, this study first locates the anomaly regions using normal samples, and then extracts CNN embeddings of the anomaly regions. During the test phase, we compute the anomaly embedding distances between query and support sets to identify the class label of query samples. Fig. \ref{Paths_comparison} compares our proposal and Prototype networks. PatchProto networks adapt the core idea of Prototype networks, one of the canonical FSL methods. Distinctively, PatchProto networks compute the prototypes in the shallow layers rather than bottleneck feature and only the embeddings of anomaly pixels (i.e., patches) are concerned. The anomaly patches are located by contrasting with normal samples, which has been widely applied in the anomaly segmentation works~\cite{DBLP:journals/corr/abs-2106-08265, 10.1007/978-3-030-68799-1_35, DBLP:journals/corr/abs-2005-02357}. 

\par In summary, we propose PatchProto networks as an effective FSL anomaly classification method, by (1) using the CNN feature embeddings of only anomaly patches to serve as the prototypes, (2) extracting features from the shallow layers. The anomaly segmentation work~\cite{DBLP:journals/corr/abs-2106-08265} has shown that using mid-level pertained network features can reduce the biases towards ImagesNet classes. In the following sections, this paper will discuss the significance of forcing model to focus on the anomaly patches. On 14 sub-datasets of MVTec-AD, PatchProto networks achieves 16\% and 10\% higher average accuracy over vanilla Prototype networks and the simple combining Prototype networks with anomaly segmentation (Prototype + segmentation).

\section{Background and related works}
\subsection{Anomaly diagnosis}
\par Many anomaly diagnosis methods ensemble both detection and segmentation facility. Most approaches learns the representations of normal data in a unsupervised manner, for which many generative models are utilized, such as GANs~\cite{10.1007/978-3-030-10925-7_1}, (Variational) AutoEncoders~\cite{10.1145/3097983.3098052, 0Variational} and flow models~\cite{Rudolph_2021_WACV}. During the test phase, the representations of novel samples are inferred and the anomaly detection results are computed by the reconstruction errors or KNN methods. For these methods, anomaly segmentation can be subsequently achieved by pixel-level reconstruction errors or some feature attribution methods.
\par Recently, some research shows that, with the help of models pretrained on large external datasets (e.g., ImageNet), the performance of industrial anomaly detection can be further improved. Approaches like SPADE~\cite{DBLP:journals/corr/abs-2005-02357}, PaDiM~\cite{10.1007/978-3-030-68799-1_35} and PatchCore~\cite{DBLP:journals/corr/abs-2106-08265} extract the image features with pretrained networks at patch-level, then use similarity measures to score the patch-level abnormality. Among them, PatchCore achieved state-of-the-art on total recall perform of both detection and segmentation. 
\par Anomaly classification is a further task that identifies the types of anomaly samples, which provides more detailed information than anomaly detection. Data scarcity is always the major issue in the anomaly classification. Natarajan et al.~\cite{7915495} propose voting-based methods to solve the over-fitting caused by small textile datasets. There are many works presented for the few-shot industrial fault classification, in the flied like rolling bearing~\cite{LI2021197}, chemical process~\cite{9329076}, and power systems~\cite{9537307}.  

\subsection{Few-shot learning}
\par FSL is a task of classification, in which the sample numbers are quite small (typically less 10 each class). In FSL setting, the labeled dataset for training is named as support set and the samples with unknown labels for testing compose the query set. A few-shot classification model uses the information from the support set to classify the query images. Given support set with $N$ different class labels and $K$ samples each class, the FSL is named $N$-way $K$-shots problem.
\par FSL is more challenging than other problems like semi-supervised or imbalanced classification, since the extremely scarce samples would easily lead to over-fitting and bad performance on test sets. The popular FSL solutions are based on meta-learning~\cite{DBLP:journals/corr/FinnAL17}, which first learn from the relevant classes of abundant samples and then transfer to minor classes. 
\par However, the meta-learning is not suitable for anomaly classification, since there is not enough relevant classes for pre-training. This paper focuses on another prevalent FSL strategy, metric learning. Its core idea is to compute the embeddings of input and use similarity metrics of embeddings to classify samples. Prototypical Networks~\cite{DBLP:journals/corr/SnellSZ17} is a typical approach. It uses a pretrained CNN to project support and query samples into the feature space, and computes the prototype for each class by averaging the embeddings of support samples. For query samples, they are classified by the distances from their embeddings to each prototype.

\par For visual anomaly diagnosis, the few-shot learning is more studied in anomaly detection. Visual Anomaly and Novelty Detection (VAND) 2023 Challenge proposes two tracks of zero-shot and few-shot anomaly detection. Given the limited number of normal samples, April-GAN~\cite{chen2023zero} and an optimized PatchCore~\cite{santos2023optimizing} achieve the best performances on detecting anomaly samples. However, the scenario that the anomaly samples are scarce (i.e., few-shot visual anomaly classification), is less concerned, while we believe this setting is more practical.

\section{Method}
\par This section present our proposal for few-shot visual anomaly classification, PatchProto networks. It consists of two major parts: anomaly prototypes and distance measures, which will be introduced in sequence. Algorithm \ref{alg} presents the pseudo code of PatchProto networks.

\begin{figure}[!t]

  \renewcommand{\algorithmicrequire}{\textbf{Input:}}
  \renewcommand{\algorithmicensure}{\textbf{Output:}}

  \begin{algorithm}[H]
    \caption{PatchProto networks}
    \begin{algorithmic}[1]
      \REQUIRE
        $\phi$: Pretrained CNN for embedding extraction
        $\mathcal{X}_N$: normal samples\\
        $\{(\mathcal{X}_S,\mathcal{Y}_S)\}$: support set \\ 
        $\mathcal{X}_Q$: query set \\
      \STATE $ \mathcal{Z}_N, \mathcal{Z}_S, \mathcal{Z}_Q = \phi(\mathcal{X}_N), \phi(\mathcal{X}_S), \phi(\mathcal{X}_Q) \vartriangleleft  $ extract embeddings 
      \STATE $s_{\mathcal{Z}_S}, s_{\mathcal{Z}_Q} = f_s(\mathcal{Z}_S , \mathcal{Z}_N), f_s(\mathcal{Z}_Q , \mathcal{Z}_N) \vartriangleleft $ compute patch-level anomaly score
      \STATE $\mathcal{Z}^P_S, \mathcal{Z}^P_Q=  f_p( \mathcal{Z}_S, s_{\mathcal{Z}_S} ), f_p(\mathcal{Z}_Q, s_{\mathcal{Z}_Q} )\vartriangleleft $ select anomaly embeddings
      \STATE $\mathcal{Z}^y = \{z^i | z^i\in \mathcal{Z}^P_S, i=y\in \mathcal{Y}\}\vartriangleleft $ gather prototypes of each class
      \STATE $d^y = D( \mathcal{Z}^P_Q, \mathcal{Z}^y), y\in \mathcal{Y}\vartriangleleft  $ measure distances to prototypes
      \ENSURE $\mathcal{Y}_Q = \{y | y = \arg\min d^y\}$ Predicted labels of query set
    \end{algorithmic}
    \label{alg}
  \end{algorithm}
\end{figure}
\subsection{Anomaly prototypes}
\par Unlike typical Prototype networks that use flattened bottleneck embeddings as the prototypes, PatchProto networks select partial mid-level embeddings that are closely related to the anomaly regions of input images. We first determine the anomaly scores of image patches by the embeddings distances to the normal samples. Specifically, we adapt scoring algorithm ($f_s$ in Algorithm \ref{alg}) of PatchCore\footnote{The anomaly scoring is not limited to one algorithm, and any pixel-wise anomaly segmentation algorithm can be applied.} that achieves state-of-the-art anomaly segmentation performance. 
\par In anomaly classification, the normal image pixels do not contain positive information to distinguish different anomaly types, and only the anomaly regions determine the image types. Hence, we need to select the embeddings that most likely correspond to the anomaly image regions. Given the embeddings $\mathcal{Z}_S$ with anomaly scores $s_{\mathcal{Z}_S}$, one direct approach is to select the top-$k$ embeddings with highest scores ($k$ is a fixed number). However, the size of anomaly region in each image is different, so a fixed embedding number is imprecise. A feasible way is to select the embeddings that contribute a certain percentage ($\gamma$) of the total score, just like the way to select PCA eigenvectors. This approach can adaptively select the most abnormal embeddings of the input images. Nevertheless, if the image anomaly region cannot be precisely located, this approach will collapse and select a very large anomaly region, which is totally misleading. To avoid this, we set a upper bound ($m$) for the anomaly patch number. Algorithm \ref{alg_sel} shows the procedure for selecting the anomaly embeddings ($f_p$ in Algorithm \ref{alg}).
\begin{figure}[!t]

  \renewcommand{\algorithmicrequire}{\textbf{Input:}}
  \renewcommand{\algorithmicensure}{\textbf{Output:}}

  \begin{algorithm}[H]
    \caption{Anomaly embedding selection (taking one image as example)}
    \begin{algorithmic}[1]
      \REQUIRE
        $z\in \mathcal{Z}$: embeddings of one image
        $s \in s_{\mathcal{Z}}$: anomaly scores of embeddings\\
        $\gamma$: threshold \\
        $m$: embedding number upper bound \\

      \STATE $z^*=\{\},\ c = 0, i=0$
      \STATE $s = softmax(s) \vartriangleleft $ normalization
      \STATE Index the elements of $z_s$ and $s$ by $s$ in decreasing order
      \WHILE{$c<\gamma$ and $|z^*|<m$}   
      \STATE $z^*= z^* \cap z[i]$
      \STATE $c += s[i] $
      \STATE $i += 1 $
      \ENDWHILE
      \ENSURE $z^*=\{\} $ Selected anomaly embeddings
    \end{algorithmic}
    \label{alg_sel}
  \end{algorithm}
\end{figure}

\par The anomaly embeddings of support and query sets are selected by the same algorithm. For support set, we simply gather the embeddings of identical anomaly type to get the prototypes $\mathcal{Z}^y$. Since the embedding sizes of support samples are not fixed, we do not take the mean of embeddings as the prototype, which is commonly applied in the other prototype-based methods.

\subsection{Distance measures}
\par Given the anomaly embeddings of support and query sets, the query samples can be classified by measuring the distances between its embeddings to the prototypes. A prototype contains a set of embeddings, so the aim is to measure the distance between two vector sets. Given a query anomaly embeddings $z_Q \in \mathcal{Z}_Q$, the distance of each single query embedding $z^{(i)}_Q$ is measured by the maximal Cosine distance w.r.t. different prototype embeddings $\mathcal{Z}^y[n]$. The distance measure of a single query sample is obtained by summation with the weights of normalized anomaly scores $s_{z^{(i)}_Q}$. Finally, we gain the class distance by the mean distance over ($N$) different prototype embeddings of the same class:
\begin{align}
  d^y = \frac{1}{N} \Sigma_{n=0}^{N}  \Sigma_{i=0}^{| z_Q|}\ s_{z^{(i)}_Q} \times \min [1- Cos(z^{(i)}_Q,\mathcal{Z}^y[n])]
\end{align}
\par Based on the class distances, we can easily predict label of query samples by selecting the closest class.

\begin{table}[t]
  \centering
    \caption{Few-shot classification accuracy on MVTec-AD dataset (averaged over subsets) (\%)}   
     \label{result_all} 
     
	\begin{threeparttable}
    \begin{tabular}{lccc}
      \toprule
                   & 1-shot & 3-shot & 5-shot     \\ \midrule
                   Matching networks~\cite{DBLP:journals/corr/HariharanG16}          & 39.0 & 42.4 & 46.0  \\
                   SimpleShot~\cite{DBLP:journals/corr/abs-1911-04623}                 & 39.6 & 48.1 & 53.9  \\
                   FineTune~\cite{DBLP:journals/corr/abs-1904-04232}                   & 39.6 & 48.3 & 54.0  \\
                   Prototype networks~\cite{DBLP:journals/corr/SnellSZ17}         & 38.3 & 49.1 & 53.7  \\
                   Segmentation\tnote{*} + Proto        &   44.2 & 54.2 & 57.6 \\
                   \textbf{PatchProto networks} (Ours)   & 57.9 & 65.5  & 68.9 \\
     \bottomrule
    \end{tabular}
  \end{threeparttable}
  \begin{tablenotes}
  \footnotesize
  \item{*} Anomaly segmentation by PatchCore.
  \end{tablenotes}
    \end{table}
\section{Experiments}
\subsection{Details}
\par The experiment is performed on the MVTec Anomaly Detection dataset. We use 14 of 15 sub-datasets\footnote{The dataset ``toothbrush'' only has one anomaly type, which is excluded.}. Each sub-dataset contains at least 200 normal samples, of which each anomaly class contains 10 to 20 samples. In experiments, we only classify anomaly samples while normal samples are not considered. We resize the images to the size of $224\times 224$ and do not apply any augmentation techniques. We use WideResNet-50-2 pretrained on ImageNet as the feature extractor, also applied for other compared methods.

\subsection{Results}
\par Table \ref{result_all} reports the average few-shot classification accuracy on MVTec-AD, which compares networks with typical FSL methods including Matching networks~\cite{DBLP:journals/corr/HariharanG16}, SimpleShot~\cite{DBLP:journals/corr/abs-1911-04623}, FineTune~\cite{DBLP:journals/corr/abs-1904-04232} and Prototype networks~\cite{DBLP:journals/corr/SnellSZ17}. Given the original anomaly images as the input, typical FSL classifiers achieve similar accuracy. When preprocessing anomaly images by segmentation, few-shot accuracy cloud be significantly improved (Segmentation + Proto). This validates that excluding irrelative image regions can effectively benefit the anomaly classification.
\par Our proposal, PatchProto networks, achieve the best few-shot classification accuracy over these methods, which is about 15\% higher than FSL methods with original input and 10\% higher than segmentation input. Intriguingly, the advantage of PatchProto networks is especially significant on 1-shot, which we believe is mainly attributed to the anomaly embedding used for prototypes.

\begin{figure*}[t]
  \centering
  \includegraphics[width=0.95\textwidth]{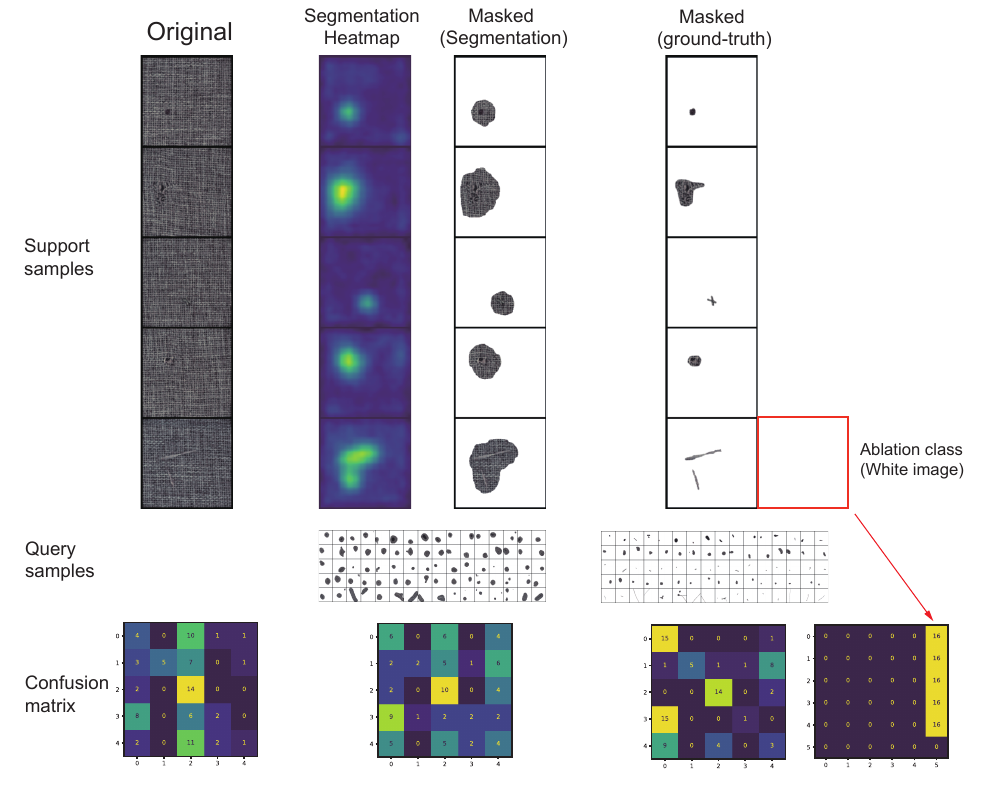}
  \caption{\textbf{Illustration of baselines for ablation study.} }
  \label{ablation}
\end{figure*}

\begin{table}[t]
  \centering
    \caption{Few-shot classification accuracy on different subsets of MVTec-AD dataset (\%)}   
     \label{origmask} 
    \begin{tabular}{lccc|ccc|ccc}
      \toprule
     &      \multicolumn{3}{c|}{PatchProto networks} &      \multicolumn{3}{c}{Masked (Segmentation)}   &    \multicolumn{3}{c}{Masked (Ground-truth)}     \\
                   & 1-shot & 3-shot & 5-shot & 1-shot & 3-shot & 5-shot  & 1-shot & 3-shot & 5-shot      \\ \midrule
                   bottle     &   52.2 & 52.0 & 57.5 & 61.8 & 69.8 & 72.9  & 73.0 & 76.5 & 77.8\\
                   cable      &   62.5 & 69.7 & 73.0 & 44.4 & 59.6 & 64.0  & 60.8 & 71.1 & 71.5\\
                   capsule    &   34.5 & 37.9 & 42.9 & 31.4 & 37.6 & 40.3  & 37.7 & 42.8 & 46.1\\
                   carpet     &   56.6 & 67.7 & 73.0 & 33.8 & 59.4 & 60.7  & 45.7 & 58.9 & 69.7\\
                   grid       &   40.6 & 38.7 & 43.2 & 36.0 & 50.0 & 48.0  & 44.4 & 53.0 & 62.7\\
                   hazelnut   &   81.0 & 95.3 & 93.7 & 50.9 & 67.1 & 77.5  & 70.0 & 82.9 & 83.3\\
                   leather    &   71.9 & 84.1 & 86.6 & 50.0 & 61.1 & 65.3  & 55.5 & 60.0 & 65.7\\
                   metal nut  &   60.8 & 69.0 & 76.3 & 54.5 & 57.4 & 60.3  & 63.8 & 78.2 & 80.9\\
                   pill       &   35.8 & 41.7 & 45.5 & 33.6 & 39.0 & 42.1  & 44.3 & 66.7 & 68.6\\
                   screw      &   38.8 & 41.5 & 42.4 & 26.2 & 29.4 & 28.0  & 57.6 & 73.4 & 74.4\\
                   tile       &   85.2 & 94.5 & 96.7 & 68.9 & 76.3 & 84.0  & 79.4 & 83.0 & 91.2\\
                   transistor &   65.6 & 78.8 & 78.6 & 66.1 & 71.4 & 83.0  & 67.8 & 79.3 & 88.0\\
                   wood       &   67.5 & 77.7 & 86.8 & 30.9 & 42.4 & 36.0  & 37.1 & 51.2 & 45.3\\
                   zipper     &   58.0 & 67.8 & 68.2 & 30.1 & 38.0 & 43.6  & 32.4 & 45.3 & 53.8\\ \midrule
                   average    &   57.9 & 65.5 & 68.9 & 44.2 & 54.2 & 57.6  & 55.0 & 65.9 & 69.9\\
     \bottomrule
    \end{tabular}
    \end{table}

\subsection{Ablation study}
\label{back}
\par It is straightforward to compare our proposal with the Prototype networks. As baselines, we process support and query sets in three different ways: (1) the original images; (2) the images masked by the PatchCore anomaly segmentation; (3) the images masked by the manually annotated (ground-truth) segmentation. Figure \ref{ablation} shows the baselines for ablation study, where the ``carpet'' subset of MVTec-AD are displayed and each anomaly class has one single support samples. 
\par For original anomaly samples, the confusion matrix of Prototype networks shows that model tends to classify samples into class ``2'' label. This is mainly due to the large normal pixels in the class ``2'' support image. Since the normal pixels take up the main body of the anomaly image and anomaly region is only a small part, it is natural for models to predict that inputs are more similar to the images that are close to normal samples (e.g., class ``2'' support image in Figure \ref{ablation}).
\par Preprocessing images with anomaly segmentation can avoid this problem, shown in the middle of Figure \ref{ablation}. Prototype networks with masked anomaly images have higher accuracy. However, the segmentation is not very precious and the masked images still have some normal regions. Hence, we test the images masked by manually annotation (ground-truth), of which the accuracy is further improved. 
\par However, the masked images will still introduce irrelative information, i.e., the white backgrounds. We add a all-white support image as the an ablation class (in the right of Figure \ref{ablation}). The confusion matrix shows that all the query samples are predicted as the most similar to the all-white image.

\par In summary, there are two major problems in the method that simply combines anomaly segmentation and few-shot learning:
\begin{itemize}
  \item Anomaly segmentation is not precious, which leads to the gap between manually annotated anomaly area (ground truth).
  \item The features of masked images are extracted by pertained networks, and the representations of the last layer are used. These embeddings represent the features of the whole masked images, not the critical anomaly regions, which introduces irrelative information.
\end{itemize}
\par This inspires us to propose PatchCore, which uses the only features of anomaly regions to calculate the prototypes.

\par Besides, Table \ref{origmask} reports the few-shot classification accuracy on different subsets. PatchProto networks achieve great improvement in most of the subsets, while ``bottle'', ``grid'', ``pill'' and ``transistor'' are exceptions. The anomaly in these subsets takes up more image regions, so the methods take the whole image as the input will perform better. 
PatchProto networks tend to focus on the most anomaly images region and ignore some global anomaly features, which causes the undesirable performance.

    \begin{figure*}[t]
      \centering
      \includegraphics[width=0.95\textwidth]{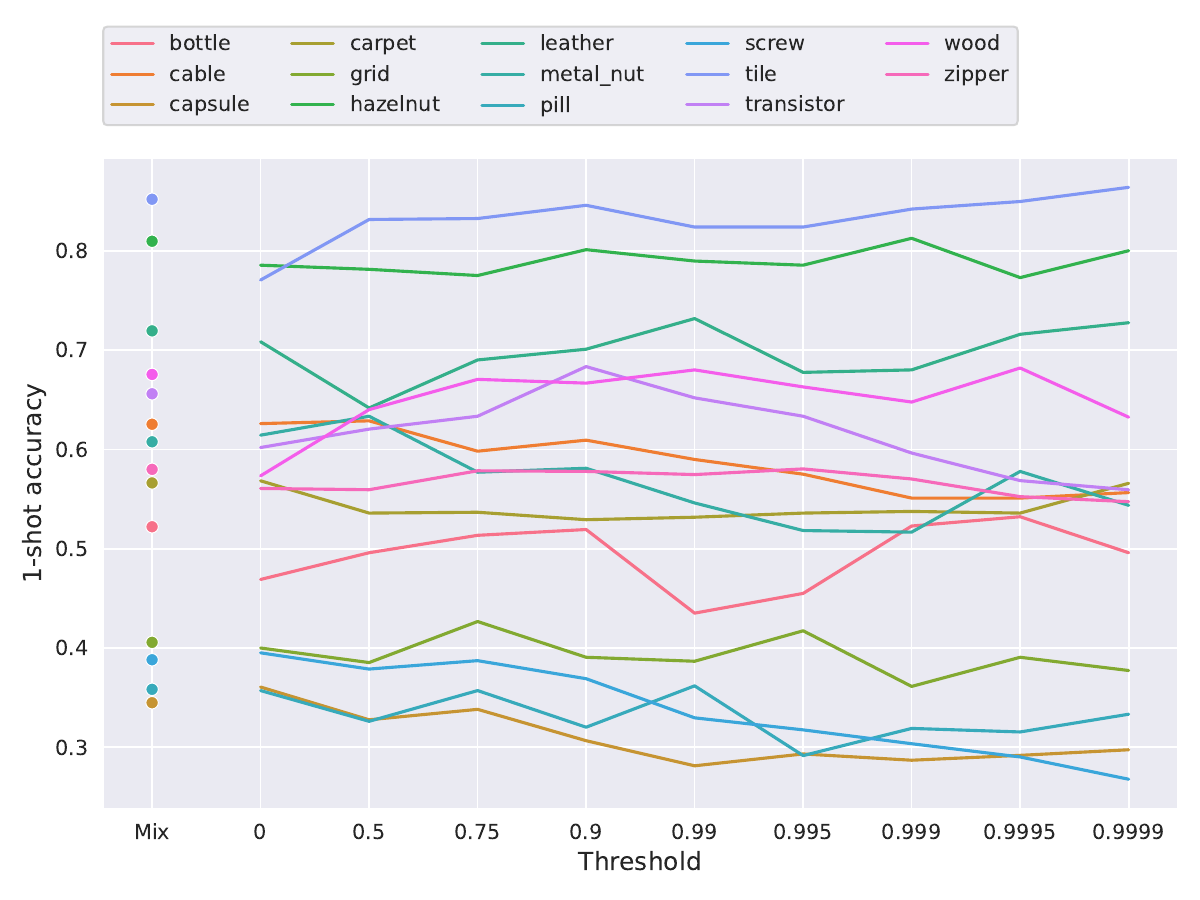}
      \caption{\textbf{1-shot classification accuracy under different threshold values.} }
      \label{1shot}
    \end{figure*}

\subsection{Parameters}   
\par The most critical parameters of PatchProto networks are thresholds to select the anomaly features, i.e., $\gamma$ in Algorithm \ref{alg_sel}. Figure~\ref{1shot} shows the 1-shot accuracy of PatchProto networks under different $\gamma$ values. Larger $\gamma$ represents that more anomaly regions are selected, while smaller $\gamma$ represents only the most anomaly regions are used (0 denotes selecting only one patch with highest anomaly score). 
\par We find that high $\gamma$ value benefits the accuracy of subsets that are easy to classify, i.e., ``tile'' and ``leather'', but some hard-classified subsets like ``capsule'' and ``screw'' achieve best performance with single anomaly patch. This is because that hard-classified subsets are also hard to segmentation and anomaly scores on them are inaccurate. The high thresholds will select large image regions, introducing irrelative information. This inspires us to set a upper bound for the number of selected anomaly features, i.e., $m$ in Algorithm \ref{alg_sel}. 
\par Shown in Figure~\ref{1shot}, this mix strategy for anomaly feature selection improves the overall accuracy.

\section{Conclusions}
This paper propose a few-shot learning method, PatchProto networks, for a practical issue that anomaly samples for training are extremely scarce. PatchProto networks are inspired by PatchCore for anomaly segmentation and Prototype networks for few-shot learning. Nevertheless, unlike classical FSL methods, PatchProto networks fully utilize the normal samples, and only extract CNN features of defective regions of interest, which serves as the prototypes for few-shot learning. The experiment results on MVTec-AD dataset show PatchProto networks significantly improve the anomaly classification accuracy under different few-shot settings. Up to now, PatchProto networks uses a simple approach to calculate the similarity between anomaly features of different length. In future, more sophisticated approach, i.e., LSTM networks, can be applied for calculating the prototype similarity.

\bibliographystyle{elsarticle-harv}
\bibliography{Bibfewshot}

\begin{thebibliography}{26}
\expandafter\ifx\csname natexlab\endcsname\relax\def\natexlab#1{#1}\fi
\expandafter\ifx\csname url\endcsname\relax
  \def\url#1{\texttt{#1}}\fi
\expandafter\ifx\csname urlprefix\endcsname\relax\def\urlprefix{URL }\fi

\bibitem[{An and Cho(2015)}]{0Variational}
An, J., Cho, S., 2015. Variational autoencoder based anomaly detection using
  reconstruction probability. Special lecture on IE 2~(1), 1--18.

\bibitem[{Bergmann et~al.(2019)Bergmann, Fauser, Sattlegger, and
  Steger}]{Bergmann_2019_CVPR}
Bergmann, P., Fauser, M., Sattlegger, D., Steger, C., June 2019. Mvtec ad -- a
  comprehensive real-world dataset for unsupervised anomaly detection. In:
  Proceedings of the IEEE/CVF Conference on Computer Vision and Pattern
  Recognition (CVPR).

\bibitem[{Chen et~al.(2019)Chen, Liu, Kira, Wang, and
  Huang}]{DBLP:journals/corr/abs-1904-04232}
Chen, W., Liu, Y., Kira, Z., Wang, Y.~F., Huang, J., 2019. A closer look at
  few-shot classification. CoRR abs/1904.04232.
\newline\urlprefix\url{http://arxiv.org/abs/1904.04232}

\bibitem[{Chen et~al.(2023)Chen, Han, and Zhang}]{chen2023zero}
Chen, X., Han, Y., Zhang, J., 2023. A zero-/few-shot anomaly classification and
  segmentation method for cvpr 2023 vand workshop challenge tracks 1\&2: 1st
  place on zero-shot ad and 4th place on few-shot ad. arXiv preprint
  arXiv:2305.17382.

\bibitem[{Cohen and Hoshen(2020)}]{DBLP:journals/corr/abs-2005-02357}
Cohen, N., Hoshen, Y., 2020. Sub-image anomaly detection with deep pyramid
  correspondences. CoRR abs/2005.02357.
\newline\urlprefix\url{https://arxiv.org/abs/2005.02357}

\bibitem[{Deecke et~al.(2019)Deecke, Vandermeulen, Ruff, Mandt, and
  Kloft}]{10.1007/978-3-030-10925-7_1}
Deecke, L., Vandermeulen, R., Ruff, L., Mandt, S., Kloft, M., 2019. Image
  anomaly detection with generative adversarial networks. In: Berlingerio, M.,
  Bonchi, F., G{\"a}rtner, T., Hurley, N., Ifrim, G. (Eds.), Machine Learning
  and Knowledge Discovery in Databases. Springer International Publishing,
  Cham, pp. 3--17.

\bibitem[{Defard et~al.(2021)Defard, Setkov, Loesch, and
  Audigier}]{10.1007/978-3-030-68799-1_35}
Defard, T., Setkov, A., Loesch, A., Audigier, R., 2021. Padim: A patch
  distribution modeling framework for anomaly detection and localization. In:
  Del~Bimbo, A., Cucchiara, R., Sclaroff, S., Farinella, G.~M., Mei, T.,
  Bertini, M., Escalante, H.~J., Vezzani, R. (Eds.), Pattern Recognition. ICPR
  International Workshops and Challenges. Springer International Publishing,
  Cham, pp. 475--489.

\bibitem[{Finn et~al.(2017)Finn, Abbeel, and
  Levine}]{DBLP:journals/corr/FinnAL17}
Finn, C., Abbeel, P., Levine, S., 2017. Model-agnostic meta-learning for fast
  adaptation of deep networks. CoRR abs/1703.03400.
\newline\urlprefix\url{http://arxiv.org/abs/1703.03400}

\bibitem[{Hariharan and Girshick(2016)}]{DBLP:journals/corr/HariharanG16}
Hariharan, B., Girshick, R.~B., 2016. Low-shot visual object recognition. CoRR
  abs/1606.02819.
\newline\urlprefix\url{http://arxiv.org/abs/1606.02819}

\bibitem[{Hu et~al.(2022)Hu, Liu, Li, Chen, and Hu}]{9537307}
Hu, Y., Liu, R., Li, X., Chen, D., Hu, Q., 2022. Task-sequencing meta learning
  for intelligent few-shot fault diagnosis with limited data. IEEE Transactions
  on Industrial Informatics 18~(6), 3894--3904.

\bibitem[{Li et~al.(2021)Li, Li, Zhang, He, Liao, and Hu}]{LI2021197}
Li, C., Li, S., Zhang, A., He, Q., Liao, Z., Hu, J., 2021. Meta-learning for
  few-shot bearing fault diagnosis under complex working conditions.
  Neurocomputing 439, 197--211.
\newline\urlprefix\url{https://www.sciencedirect.com/science/article/pii/S0925231221001818}

\bibitem[{Ma et~al.(2021)Ma, Wu, Xue, Yang, Zhou, Sheng, Xiong, and
  Akoglu}]{ma2021comprehensive}
Ma, X., Wu, J., Xue, S., Yang, J., Zhou, C., Sheng, Q.~Z., Xiong, H., Akoglu,
  L., 2021. A comprehensive survey on graph anomaly detection with deep
  learning. IEEE Transactions on Knowledge and Data Engineering.

\bibitem[{Natarajan et~al.(2017)Natarajan, Hung, Vaikundam, and Chia}]{7915495}
Natarajan, V., Hung, T.-Y., Vaikundam, S., Chia, L.-T., 2017. Convolutional
  networks for voting-based anomaly classification in metal surface inspection.
  In: 2017 IEEE International Conference on Industrial Technology (ICIT). pp.
  986--991.

\bibitem[{Pang et~al.(2021)Pang, Shen, Cao, and Hengel}]{10.1145/3439950}
Pang, G., Shen, C., Cao, L., Hengel, A. V.~D., mar 2021. Deep learning for
  anomaly detection: A review. ACM Comput. Surv. 54~(2).
\newline\urlprefix\url{https://doi.org/10.1145/3439950}

\bibitem[{Roth et~al.(2021)Roth, Pemula, Zepeda, Sch{\"{o}}lkopf, Brox, and
  Gehler}]{DBLP:journals/corr/abs-2106-08265}
Roth, K., Pemula, L., Zepeda, J., Sch{\"{o}}lkopf, B., Brox, T., Gehler, P.~V.,
  2021. Towards total recall in industrial anomaly detection. CoRR
  abs/2106.08265.
\newline\urlprefix\url{https://arxiv.org/abs/2106.08265}

\bibitem[{Rudolph et~al.(2021)Rudolph, Wandt, and
  Rosenhahn}]{Rudolph_2021_WACV}
Rudolph, M., Wandt, B., Rosenhahn, B., January 2021. Same same but differnet:
  Semi-supervised defect detection with normalizing flows. In: Proceedings of
  the IEEE/CVF Winter Conference on Applications of Computer Vision (WACV). pp.
  1907--1916.

\bibitem[{Santos et~al.(2023)Santos, Tran, and Rippel}]{santos2023optimizing}
Santos, J., Tran, T., Rippel, O., 2023. Optimizing patchcore for few/many-shot
  anomaly detection.

\bibitem[{Shi et~al.(2021)Shi, Yang, and Qi}]{SHI20219}
Shi, Y., Yang, J., Qi, Z., 2021. Unsupervised anomaly segmentation via deep
  feature reconstruction. Neurocomputing 424, 9--22.
\newline\urlprefix\url{https://www.sciencedirect.com/science/article/pii/S0925231220317951}

\bibitem[{Simon(????)}]{Omniglot}
Simon, A., ???? Omniglot - writing systems and languages of the world. 25th
  September 2023. www.omniglot.com.

\bibitem[{Snell et~al.(2017)Snell, Swersky, and
  Zemel}]{DBLP:journals/corr/SnellSZ17}
Snell, J., Swersky, K., Zemel, R.~S., 2017. Prototypical networks for few-shot
  learning. CoRR abs/1703.05175.
\newline\urlprefix\url{http://arxiv.org/abs/1703.05175}

\bibitem[{Vinyals et~al.(2016)Vinyals, Blundell, Lillicrap, Kavukcuoglu, and
  Wierstra}]{DBLP:journals/corr/VinyalsBLKW16}
Vinyals, O., Blundell, C., Lillicrap, T.~P., Kavukcuoglu, K., Wierstra, D.,
  2016. Matching networks for one shot learning. CoRR abs/1606.04080.
\newline\urlprefix\url{http://arxiv.org/abs/1606.04080}

\bibitem[{Wah et~al.(2011)Wah, Branson, Welinder, Perona, and
  Belongie}]{2011The}
Wah, C., Branson, S., Welinder, P., Perona, P., Belongie, S., 2011. The
  caltech-ucsd birds-200-2011 dataset. california institute of technology.

\bibitem[{Wang et~al.(2019)Wang, Chao, Weinberger, and van~der
  Maaten}]{DBLP:journals/corr/abs-1911-04623}
Wang, Y., Chao, W., Weinberger, K.~Q., van~der Maaten, L., 2019. Simpleshot:
  Revisiting nearest-neighbor classification for few-shot learning. CoRR
  abs/1911.04623.
\newline\urlprefix\url{http://arxiv.org/abs/1911.04623}

\bibitem[{Yi and Yoon(2020)}]{Yi_2020_ACCV}
Yi, J., Yoon, S., November 2020. Patch svdd: Patch-level svdd for anomaly
  detection and segmentation. In: Proceedings of the Asian Conference on
  Computer Vision (ACCV).

\bibitem[{Zhou and Paffenroth(2017)}]{10.1145/3097983.3098052}
Zhou, C., Paffenroth, R.~C., 2017. Anomaly detection with robust deep
  autoencoders. In: Proceedings of the 23rd ACM SIGKDD International Conference
  on Knowledge Discovery and Data Mining. KDD '17. Association for Computing
  Machinery, New York, NY, USA, p. 665–674.
\newline\urlprefix\url{https://doi.org/10.1145/3097983.3098052}

\bibitem[{Zhuo and Ge(2021)}]{9329076}
Zhuo, Y., Ge, Z., 2021. Auxiliary information-guided industrial data
  augmentation for any-shot fault learning and diagnosis. IEEE Transactions on
  Industrial Informatics 17~(11), 7535--7545.

\end{thebibliography}

\end{document}